\documentclass{article}

\usepackage[preprint, nonatbib]{neurips_2023}

\usepackage[utf8]{inputenc} % allow utf-8 input
\usepackage[T1]{fontenc}    % use 8-bit T1 fonts
\usepackage[hidelinks]{hyperref}       % hyperlinks
\usepackage{url}            % simple URL typesetting
\usepackage{booktabs}       % professional-quality tables
\usepackage{amsfonts}       % blackboard math symbols
\usepackage{nicefrac}       % compact symbols for 1/2, etc.
\usepackage{microtype}      % microtypography
\usepackage{xcolor}         % colors
\usepackage{graphicx}
\usepackage{mleftright}
\usepackage{multirow}
\usepackage{subcaption,soul}
\usepackage{csvsimple}
\usepackage{caption}
\usepackage[numbers,sort&compress,sectionbib]{natbib}
\newcommand{\round}[1]{\ensuremath{\lfloor#1\rceil}}
\usepackage{comment}
\usepackage{amsmath}

\title{Starting Positions Matter: A Study on Better Weight Initialization for Neural Network Quantization}

\author{%
  Stone Yun\\
  Vision and Image Processing Research Group, University of Waterloo\\
  \texttt{s22yun@uwaterloo.ca}\\
  \And
  Alexander Wong\\
  Vision and Image Processing Research Group, University of Waterloo\\
  \texttt{a28wong@uwaterloo.ca} \\
}

\begin{document}

\maketitle

\begin{abstract}
  Deep neural network (DNN) quantization for fast, efficient inference has been an important tool in limiting the cost of machine learning (ML) model inference. Quantization-specific model development techniques such as regularization, quantization-aware training, and quantization-robustness penalties have served to greatly boost the accuracy and robustness of modern DNNs. However, very little exploration has been done on improving the initial conditions of DNN training for quantization. Just as random weight initialization has been shown to significantly impact test accuracy of floating point models, it would make sense that different weight initialization methods impact quantization robustness of trained models. We present an extensive study examining the effects of different weight initializations on a variety of CNN building blocks commonly used in efficient CNNs. This analysis reveals that even with varying CNN architectures, the choice of random weight initializer can significantly affect final quantization robustness. Next, we explore a new method for quantization-robust CNN initialization~---~using Graph Hypernetworks (GHN) to predict parameters of quantized DNNs. Besides showing that GHN-predicted parameters are quantization-robust after regular float32 pretraining (of the GHN), we find that finetuning GHNs to predict parameters for quantized graphs (which we call GHN-QAT) can further improve quantized accuracy of CNNs. Notably, GHN-QAT shows significant accuracy improvements for even 4-bit quantization and better-than-random accuracy for 2-bits. To the best of our knowledge, this is the first in-depth study on quantization-aware DNN weight initialization. GHN-QAT offers a novel approach to quantized DNN model design. Future investigations, such as using GHN-QAT-initialized parameters for quantization-aware training, can further streamline the DNN quantization process.
\end{abstract}

\section{Introduction}

Besides exponentially increasing modern massive models, there are many use-cases with a need for resource efficient deep neural networks (DNN). For example, embedded systems and other mobile AI computing requires fast, efficient DNNs that fit tight budgets on computation, latency, and power. In terms of efficient inference, developing DNNs for resource constrained AI has led to several different approaches to tackle the efficiency-accuracy tradeoff. These include fast, computationally efficient DNN architectures~\cite{MobileNetV1, MobileNetV2, FactorizeNet} and limited precision quantization for low-power inference~\cite{Nagel_DFQ, TFQuantize, LogQuant}. In resource constrained AI, combining computationally efficient architectures with fixed-precision, integer quantization such as in~\cite{TFQuantize} has become an essential tool for deploying fast, efficient convolutional neural networks (CNN) to the edge. However, perturbations induced by quantization of weights and activations can change DNN behaviour in non-trivial ways and developing efficient models for low-power, quantized inference while retaining close to original floating point accuracy is a very challenging task. In some cases, state-of-the-art performance can have significant degradation after quantization of weights and activations~\cite{MobileNetsQuantizePoorly}. 

Despite significant works demonstrating ways to recover near floating-point performance using reduced precision inference, there is still limited understanding of how different design decisions can affect the quantized inference behaviour of CNNs. For example, random weights initialization strategies are often designed with the goal of solving issues such as vanishing/exploding gradient \cite{GlorotInit, HeInit, hanin2018}. However, an often-overlooked aspect of weights initialization is its impact on the final trained distributions of each layer. As they determine our starting point on the loss surface, initial distributions of each weight tensor will significantly impact the final trained distributions of a model. With regards to quantization, this means that weight initialization choices will impact the dynamic ranges of the weights and activations in a trained CNN as well as other aspects of the learned distributions. Thus, affecting the noise in our system and the expected quantized inference behaviour.

Besides randomly initialized neural networks, Graph Hypernetworks (GHN) have also become an exciting new method for initializing DNNs with higher accuracy. Recent works by~\cite{PPUDA}~and~\cite{GHN1} have shown remarkable accuracy using GHNs to predict all trainable parameters of unseen DNNs in a \textit{single forward pass}. For example, \cite{PPUDA} report that their GHN-2 can predict the parameters of an unseen ResNet-50 to achieve 60\% accuracy on CIFAR-10. Thus, saving thousands of training steps compared to random initialization plus iterative optimization. We adapt GHNs for quantization in our investigation and further explore how GHNs can be used for better initialization of quantized CNNs. While GHN-predicted parameters are still significantly less accurate than iteratively optimized models, the high accuracy of GHN-predicted parameters could potentially make for an excellent parameter initialization that significantly amortizes a couple thousand steps of training. Previous works~\cite{PPUDA, GHN1, GHN3} have mainly focused on the use of GHNs for floating point models. By contrast, we explore the use of GHNs for initializing efficient \textbf{quantized} CNNs and generate a corresponding design space for GHN finetuning.

First, we present an in-depth study on the effects of random initialization on quantized accuracy of varying CNN architectures. We show that there are significant differences in final trained, quantized accuracy for different initialization methods. For this random initialization study, we follow the methods of ~\cite{stone_masc_thesis} and perform some layerwise analysis of the trained model distributions. Thus, observing how different initializations can lead to varying ranges and distributions and in turn, varying quantized accuracy. This observation naturally leads us to ask, are there better ways to initialize CNN parameters for training quantization-robust models?

To answer this, we present a novel exploration of the use of GHNs to predict quantization-robust efficient CNN parameters (which we will generally refer to as GHN-Q). By finetuning GHNs on a mobile-friendly CNN architecture space, we explore the use of adapting GHNs specifically to target efficient, low-power quantized CNNs. We find that the quantized accuracy of GHN-predicted parameters is surprisingly robust after float32 finetuning on the target design space and even prior to any additional quantization training (of the GHN). Finally, we demonstrate that quantization-aware training of GHNs (which we call GHN-QAT) can significantly improve quantized accuracy of low bitwidth networks such as 4-bit weights and 4-bit activations, and even achieve better-than-random accuracy for 2-bit weights/2-bit activation CNNs. Thus, we contribute a novel, in-depth study of GHNs for quantization-robust initialization of low-precision, efficient CNN architectures.

\section{Background and Related Works}
\subsection{Fixed point quantization for CNNs}
Various works have explored different quantization algorithms~\cite{TFQuantize, LogQuant, VectorQuant} to minimize the loss of information when mapping the weights and activations of a CNN into a discretized space. With these methods, the weights and activations can be represented as an n-bit (most often 8-bit) integer rather than 32-bit floating point numbers. Consequently, simple integer multiply-accumulate (MAC) operations can be performed rather than costly floating point arithmetic. This leads to significant savings in both computation (integer arithmetic) and storage (typically 8-bits per value or less).

Mobile hardware accelerators are usually limited in the types of operations that can be massively parallelized for fast execution. Thus, more complex quantization methods are often not supported by existing hardware. As such, other works have focused on quantization-algorithm-specific optimization methods (e.g., targeting 8-bit uniform quantization). These include quantization-aware fine-tuning \cite{TFQuantize} and differential optimization of quantization parameters \cite{TQT, PACT}, e.g., finding the optimal max/min thresholds of each weight/activation tensor for minimal quantized degradation. These methods train a model that is robust to quantized perturbations by simulating the error/noise of fixed point arithmetic.

In this study, we focus on using GHNs to predict efficient CNN models for existing uniform quantization methods. However, as we will see later, GHNs can be flexibly adapted to various quantization bitwidths and thus a natural extension could include different quantization schemes as well (e.g., LogQuant). The linear, uniform quantization equation is described in Eq.~\ref{eq:quantize_fn} where a quantized, integer representation $Q$ is obtained from its corresponding real number, $R$ using an affine/linear mapping followed by the necessary rounding to the nearest integer (denoted by $\round{\cdot}$). The quantization scaling factor $s$ (or alternatively, the ``stepsize'') and quantization zeropoint $Z$ (or ``offset'') are defined in Eqs.~\ref{eq:quant_s} and~\ref{eq:quant_Z} respectively where $N$ is the number of bits used for the integer representation. The values $min, max$ are obtained either from the absolute min/max ranges of the tensor to-be-quantized or they are defined by some other method of computing the quantization range (such as percentile-clipping).

\begin{equation}
\label{eq:quantize_fn}
Q = \round{\frac{clamp(R, min, max)}{s}} + Z\
\end{equation}

\begin{equation}
\label{eq:quant_s}
s = \frac{(max - min)}{2^{N} - 1}\
\end{equation}

\begin{equation}
\label{eq:quant_Z}
Z = \round{\frac{0 - min}{s}}
\end{equation}

Also important to note is that for CNNs with Batch Normalization (BatchNorm) \cite{BatchNorm}, best practices for optimal latency is to fold the BatchNorm into the convolutional weights of the preceding Conv2d operation and then quantize the resulting weight tensor~\cite{TFQuantize} (i.e., BatchNorm-Folded/BNFolded weights) such as in Equation~\ref{eq:bn_fold} where $\gamma$ is a learnable parameter, $\sigma^2_B$ is batch variance, $EMA()$ is an exponential moving average, and $\epsilon$ is a small constant. This enables the deployed model to be simply a Conv-Relu block rather than Conv-Batchnorm-Relu which can lead to considerable speedup depending on the underlying hardware and software implementations. Thus, when modelling quantization in CNNs with BatchNorm, the quantization needs to be simulated on the BatchNorm-Folded weights if those are the tensors-to-be-quantized.

\begin{equation}
\label{eq:bn_fold}
w_{bnfold} = \frac{\gamma w}{\sqrt{EMA(\sigma^2_B) + \epsilon}}
\vspace{-0.1in}
\end{equation}

\subsection{Modelling Quantization Errors in CNNs}
\label{sec:model_quant_err}
There are two main ways to model the effects and errors induced by neural network quantization: quantization simulation (SimQuant) such as that introduced in~\cite{TFQuantize} and additive noise modelling (NoiseQuant) like in~\cite{NICE, DiffQ_PseudoQuantNoise}. SimQuant inserts Quant and Dequant nodes into the CNN graph to directly simulate the errors induced by quantization through rounding and clipping of weights and activations. This method enables directly simulating the effects of quantization on a CNN forward-pass. However, the gradient of the rounding operation is zero almost everywhere and must be estimated somehow. A simple, and surprisingly effective way is to use the straight-through estimator (STE) such as in~\cite{TFQuantize} where we assume that the rounding operation's gradient is an identity function.

NoiseQuant assumes a random noise model for the quantization error and models the effects of quantization as additive uniform noise (see $NoiseQ(x)$, Eq.~\ref{eq:quant_noise}). Here, we sample quantization noise from a uniform distribution proportional to the quantization stepsize $\Delta$ and add it to the weights and activations tensors. Alternatively, if we want to train the stepsize, the sampling disribution could also be defined as in Eq.~\ref{eq:quant_noise2}. NoiseQuant can be preferred for computing exact gradients of weights and activations without using STE. However, NoiseQuant is often a very slow model since sampling the NoiseQuant tensors double memory requirements. Thus, a viable middle-ground could involve using NoiseQuant for weights and SimQuant for activations.

\begin{equation}
\label{eq:quant_noise}
NoiseQ(x) = clamp(x, a, b) + \epsilon; \text{ }   \epsilon \sim \mathcal{U}[-\Delta/2, \Delta/2]
\end{equation}

\begin{equation}
\label{eq:quant_noise2}
\epsilon \sim \frac{\Delta}{2} \cdot \mathcal{U}[-1, 1]
\end{equation}

\subsection{Random Weights Initialization}
\label{subsec:random_init_bg}
In early research, neural network parameters were often randomly initialized based on sampling from a single normal or uniform distribution. The respective variance and range of these distributions would be hyperparameters for the practitioner to decide. Several works such as \cite{GlorotInit, HeInit, hanin2018} have shown how more intelligent, ``layer/channel-aware'' random weights initialization strategies can solve issues of vanishing and exploding gradients. These works define \textit{fan\_in} and \textit{fan\_out} of a fully connected layer as the input/output units respectively. For convolution, it is defined as Eq.~\ref{eq:fan_inout} where \textit{K} is the kernel width (assume square kernel). They provide mathematical proofs on their proposed fan\_in/fan\_out-aware initialization strategies that scale the variance of gradients at each layer. Thus, avoiding failure modes created by vanishing and exploding gradients. While the introduction of BatchNorm layers has greatly mitigated training issues involving gradient scales \cite{BatchNorm}, the choice of ``where to begin'' in the parameterized loss space is still extremely relevant. An often-overlooked effect of these initialization strategies is their impact on the trained dynamic ranges of each layer. As gradient descent is a noisy, iterative process with small, incremental steps, the final dynamic ranges of each layer are profoundly impacted by their starting point.

\begin{equation}
\label{eq:fan_inout}
fan_{in/out} = K \times K \times channel_{in/out}\
\vspace{-0.05in}
\end{equation}

\subsection{Graph Hypernetworks and Parameter Prediction}
One exciting new area of research on CNN parameter initialization is neural network parameter prediction using hypernetworks. First proposed in~\cite{HyperNetworks}, hypernetworks learn to predict the optimal parameters for each layer of a network with respect to some learning task. However, the original hypernetworks needed to be retrained for each new network and thus, could not be extended/scaled to unseen networks (that is, network architectures that the hypernetwork has not been optimized on). As an alternative, recent works by~\cite{PPUDA}~and~\cite{GHN1} have explored using Graph Hypernetworks (GHN) to predict all trainable parameters of \textbf{unseen} DNNs. In GHNs, DNNs are represented as a graph where each operation/layer is a node encoded with information about the operation (e.g., convolution, 3x3 filter, 64 channels), and the connections of a given node to other layers are encoded as edges. GHNs process this graph representation using message passing\cite{scarselli_gnn, li_gatedgnn, gilmer_messagegnn, sanchez_distillpub_gnn} wherein each given node is updated by processing and aggregating features of adjacent nodes together with features of the current node. Thus, GHNs can analyze the topology of various DNN architectures and learn representations for various DNN connection patterns that can be used for predicting performant parameters of unseen network architectures.

\begin{figure*}
\centerline{\includegraphics[width=13cm]{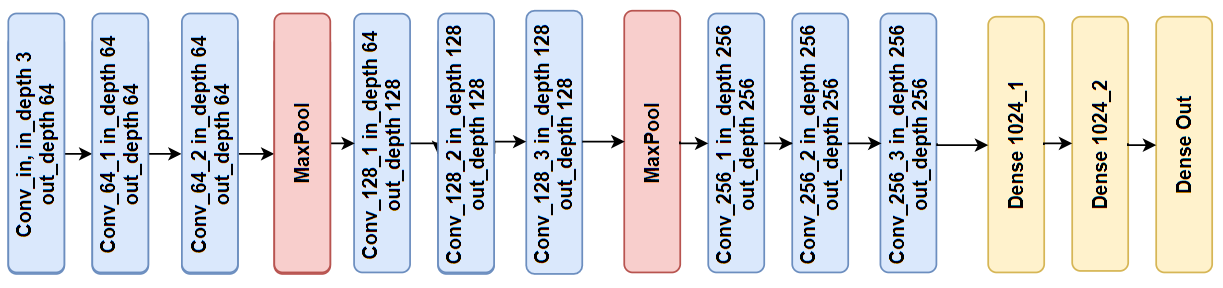}}
\caption{\footnotesize{}\textbf{General Macroarchitecture of the CNN}. For our analysis we use a fixed macro-architecture so that we can isolate the interactions between various weight initialization strategies and a few different convolutional layer choices. We train four variations of this macro-architecture determined by the type of conv-block used at each layer: Regular\_Conv\_With\_BN, Regular\_Conv\_No\_BN, DWS\_Conv\_With\_BN, and DWS\_Conv\_No\_BN. These respectively correspond to using regular convolution followed by BatchNorm and Relu, regular convolution followed by only Relu and no BatchNorm, depthwise separable (DWS) convolution blocks with BatchNorm and Relu after each convolution layer (same as the MobileNets block in \cite{MobileNet}), and finally depthwise seperable convolution with only Relu and \textbf{no} BatchNorm after each convolution layer.  The very first convolution layer stays fixed for all architectures, but follows the With/Without BatchNorm behaviour of the rest of the layers.}
\label{fig:factorizenet}
\end{figure*}

\section{Random Initialization Study}
\label{sec:random_inits_experiment}
For our initial experiment we use a simple, VGG-like macroarchitecture with four variations that differ in the micro-architecture of each layer (e.g., type of convolution block used, use of BatchNorm and Relu etc. See Figure~\ref{fig:factorizenet} for the general macro-architecture and details on the different variations of convolution layers). Our four CNNs are trained and tested on CIFAR-10 with a wide variety of different weight initialization strategies sampling from both uniform (Uni/RandUni) and normal (Norm/RandNorm) distributions. With considerations of dynamic range in mind, we seek to select distributions for the naive methods that would roughly correspond to small, medium, and large initial weights ranges. For uniform initialization this respectively corresponds to distributions of \{$\mathcal{U}[-0.25, 0.25]$, $\mathcal{U}[-0.5, 0.5]$, $\mathcal{U}[-1, 1]$\}. For normal/Gaussian initialization this respectively corresponds to distributions of \{$\mathcal{N}(0, 0.1^2)$, $\mathcal{N}(0, 0.5^2)$, $\mathcal{N}(0, 1)$\}. As mentioned in~\ref{subsec:random_init_bg}, there are also more intelligent, ``layer-aware'' initialization strategies. For the layer-aware initializations, we use four commonly used methods introduced in \cite{GlorotInit, HeInit}. Named after the authors, we call them Glorot Uniform (GlorotUni) and Glorot Normal (GlorotNorm) from \cite{GlorotInit}, He Uniform (HeUni) and He Normal (HeNorm) from \cite{HeInit}. In these works, the distribution range (for uniform sampling) and standard deviation (for normal sampling) for each layer are calculated based on \textit{fan\_in}, \textit{fan\_out}, or some combination of the two. We choose to focus on only the convolution layers and so the fully connected layers are always initialized using Glorot Uniform initialization. Furthermore, we also keep the weight initialization of the first convolution layer constant; only Glorot Uniform initialization is used. This was to keep the input convolution layer as constant as possible. Furthermore, all Dense/Linear layers are followed by Dropout with a dropping probability of 50\%.

Based on initial results showing Glorot Uniform having the most success in 32-bit floating point (fp32 or float32) accuracy, we further experiment with Modified Glorot Uniform (ModGlorotUni) weights initialization strategies. The method of computing the max/min range of the uniform sampling distribution in Glorot Uniform initialization can be generalized as Eq.~\ref{eq:glorot_uni}. In the original paper, \textit{C = 6}. We select two values of C that would roughly correspond to medium and large ranges ($C = 36$ and $C = 1296$). The original Glorot Uniform leads to fairly small ranges. 

\begin{equation}
\label{eq:glorot_uni}
max/min = \pm \sqrt{\frac{C}{fan\_in + fan\_out}}
\end{equation}

\begin{table*}
	\caption{Detailed results for each combination of weight initialization strategy and CNN architecture. The initialization strategies that suffered from vanishing/exploding gradients are ommitted. DWS\_Conv\_No\_BN\_GlorotUni kept for illustrative purposes.}
	\setlength{\tabcolsep}{4pt}
	\renewcommand{\arraystretch}{1}
	\centering
	\scalebox{0.77}{
		\csvreader[tabular=|l|c|c|c|c|c|c|,
        table head=\hline Network Architecture & FP32 Accuracy & QUINT8 Accuracy & QMSE & QCE & Percent  Accuracy Decrease\\\hline,
        late after line=\\\hline]%
        {frozen_quant_results.csv}{Network Architecture=\Network, FP32 Accuracy=\FP32, QUINT8 Accuracy=\QUINT8, QMSE=\QMSE, QCE=\QCE, Percent Accuracy Decrease=\Percent}%
        {\Network & \FP32 & \QUINT8 & \QMSE & \QCE & \Percent}%
		}
		\label{tab:simple_block_result}
		
\end{table*}

\begin{figure*}[ht]
\centerline{\includegraphics[width=13cm]{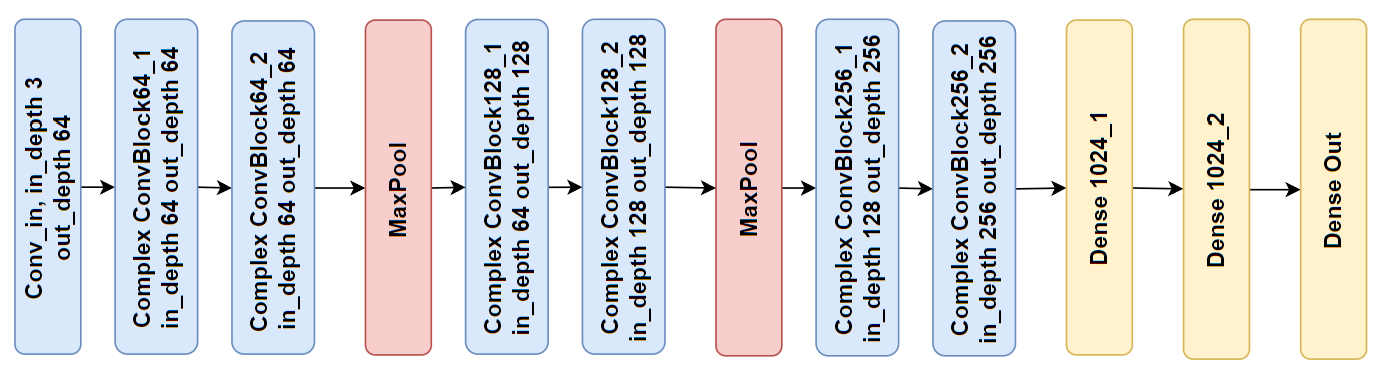}}
\caption{\footnotesize{}\textbf{General Macroarchitecture of the ``Complex Block'' CNN}. Similar to the first study, we fix the macro-architecture and then vary the type of Complex Block used (i.e., Basic Residual vs. Inverted Bottleneck Residual). The very first convolution layer stays fixed for all architectures, but follows the With/Without BatchNorm behaviour of the rest of the layers.}
\label{fig:complexblknet}
\end{figure*}

To further investigate the impact of random initialization on quantized accuracy, we repeat our ablation study with a set of more complex ConvBlock units (see Fig.~\ref{fig:complexblknet}), namely basic residual blocks such as in~\cite{ResNet} and inverted bottleneck residual blocks described in MobileNet-V2~\cite{MobileNetV2}. Both block types have more complex structure and are widely used in state-of-the-art vision tasks. The MobileNet-V2 blocks are especially of interest since they are one of the most common building blocks to use for efficient on-device CNNs.

\subsection{Experimental Setup}
Each network is trained for 200 epochs of SGD with Momentum~=~0.9 and batch-size~=~128. Initial learning rate is 0.01 and we scale it by 0.1 at the 75th, 120th, and 170th epochs. For the activation range tracking we perform top/bottom 1\% clipping computed on a random sample of 1024 training samples. Basic data augmentation includes vertical/horizontal shift, zoom, vertical/horizontal flip and rotation. We use Tensorflow for training and quantizing the weights and activations to quantized 8-bit unsigned integer (quint8) format.

For each network we evaluate testing performance with respect to 4 metrics: fp32 accuracy, quint8 accuracy, quantized mean-squared error (QMSE), and quantized crossentropy (QCE). Results for the simple-conv-block study are presented in Table~\ref{tab:simple_block_result} and complex-conv-block results are in Table~\ref{tab:complex_block_result}. QMSE refers to the MSE between the fp32 network outputs and the quint8 network outputs after dequantization. Similarly, QCE measures the cross entropy between the fp32 network outputs and the dequantized quint8 network outputs. While QMSE directly measures how much the quint8 network outputs deviate from the fp32 network, QCE quantifies the difference in the distribution of the network outputs. For classification tasks, the quantized network can predict the same class as the fp32 network, despite deviations in logit values, if the overall shape of the output distribution is similar. Therefore QCE can sometimes be more reflective of differences in quantized behaviour. Additionally, we also observe the percent accuracy degradation (change in accuracy divided by fp32 accuracy) of each network after quantization. Though these quantities often track together, there can be scenarios where a network with more QMSE or QCE actually has less relative quantization degradation from a pure accuracy standpoint. This is likely explained by favourable rounding within the network leading to predictions not being ``flipped'' by quantization.

\begin{table*}
	\caption{Effect of weight initialization on quantized accuracy of more complex blocks. The initialization strategies that suffered from vanishing/exploding gradients are ommitted.}
	\setlength{\tabcolsep}{4pt}
	\renewcommand{\arraystretch}{1}
	\centering
	\scalebox{0.77}{
		\csvreader[tabular=|l|c|c|c|c|c|c|,
        table head=\hline Network Architecture & FP32 Accuracy & QUINT8 Accuracy & QMSE & QCE & Percent  Accuracy Decrease\\\hline,
        late after line=\\\hline]%
        {frozen_quant_results2.csv}{Network Architecture=\Network, FP32 Accuracy=\FP32, QUINT8 Accuracy=\QUINT8, QMSE=\QMSE, QCE=\QCE, Percent Accuracy Decrease=\Percent}%
        {\Network & \FP32 & \QUINT8 & \QMSE & \QCE & \Percent}%
		}
		\label{tab:complex_block_result}
		
\end{table*}

\begin{figure}
\vspace{-0.15in}
    \centering
    \begin{minipage}{0.5\textwidth}
        \centering
        \includegraphics[width=0.97\textwidth]{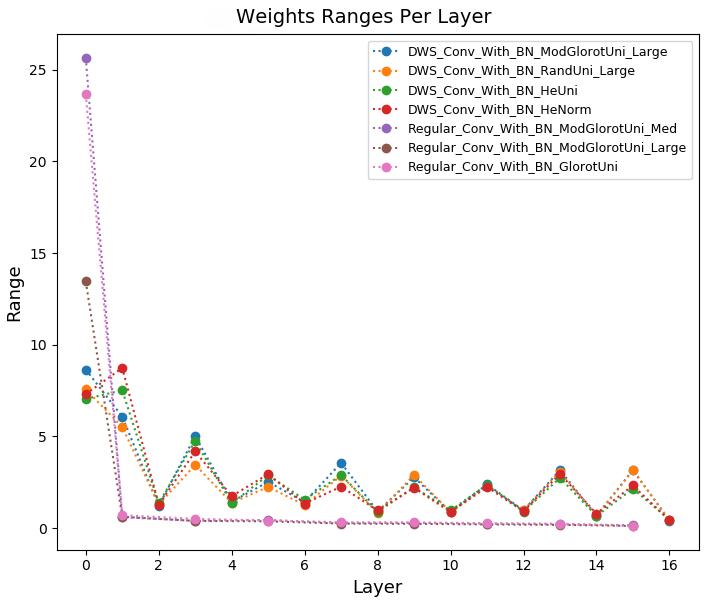} % second figure itself
    \end{minipage}\hfill
    \begin{minipage}{0.48\textwidth}
        \centering
        \includegraphics[width=0.96\textwidth]{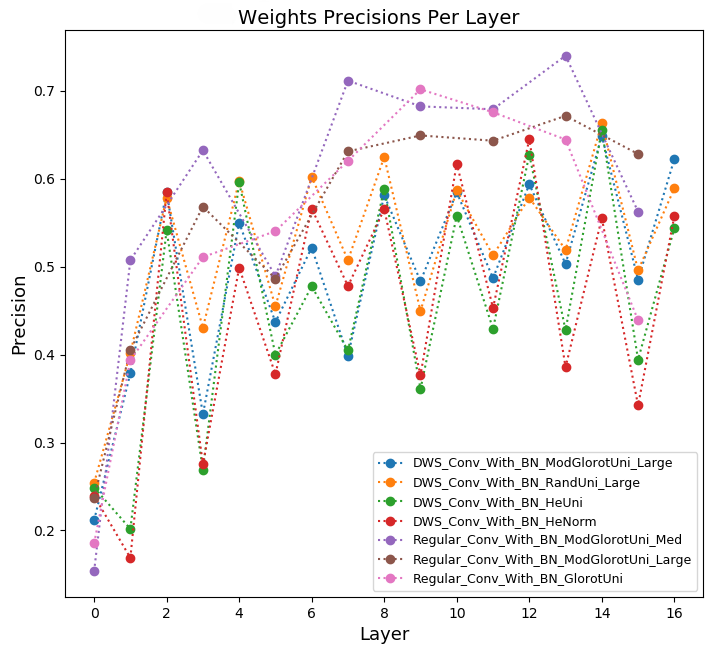} % second figure itself
    \end{minipage}\hfill
    \begin{minipage}{0.5\textwidth}
        \centering
        \includegraphics[width=0.97\textwidth]{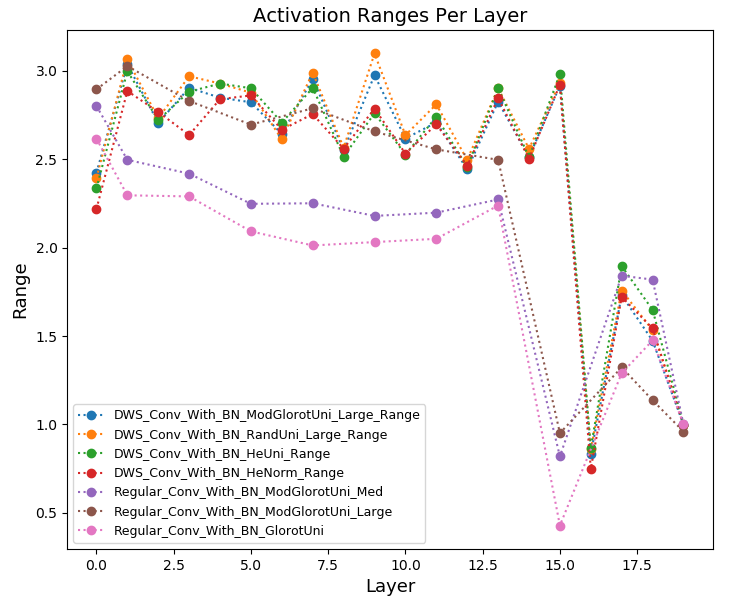} % second figure itself
    \end{minipage}\hfill
    \begin{minipage}{0.48\textwidth}
        \centering
        \includegraphics[width=0.96\textwidth]{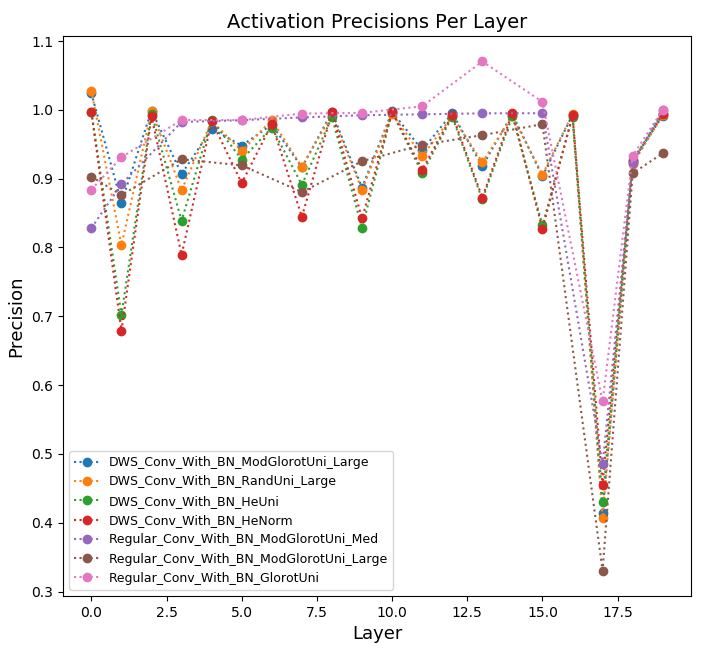} % second figure itself
    \end{minipage}
    \vspace{-0.07in}
    \caption{\footnotesize{} Layerwise plots of selected Simple-Conv-Block networks. Includes both Regular-Conv and DWS-Conv. All Weights plots are using BN-Folded Weights}
    \vspace{-0.22in}
\label{fig:focus-analysis}
\end{figure}

\begin{figure}
\vspace{-0.15in}
    \centering
    \begin{minipage}{0.5\textwidth}
        \centering
        \includegraphics[width=0.98\textwidth]{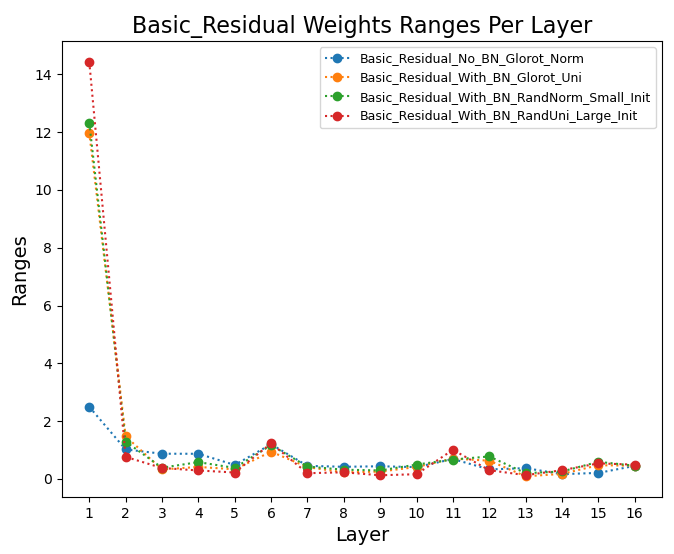} % first figure itself
    \end{minipage}\hfill
    \begin{minipage}{0.5\textwidth}
        \centering
        \includegraphics[width=0.96\textwidth]{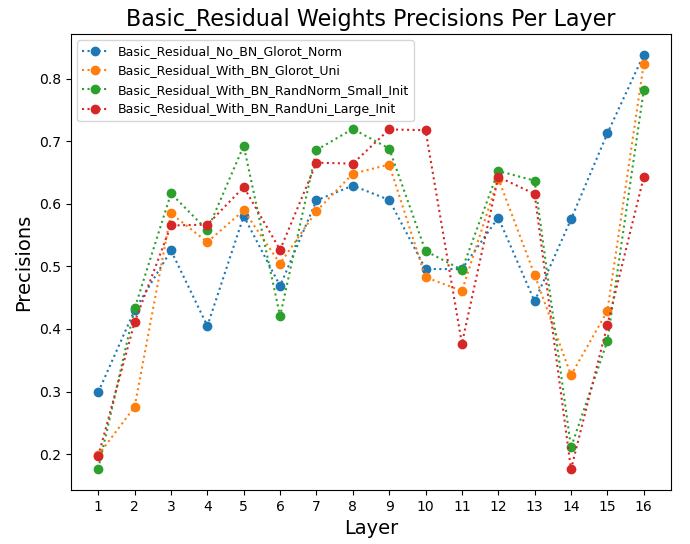} % second figure itself
    \end{minipage}\hfill
    \begin{minipage}{0.5\textwidth}
        \centering
        \includegraphics[width=0.98\textwidth]{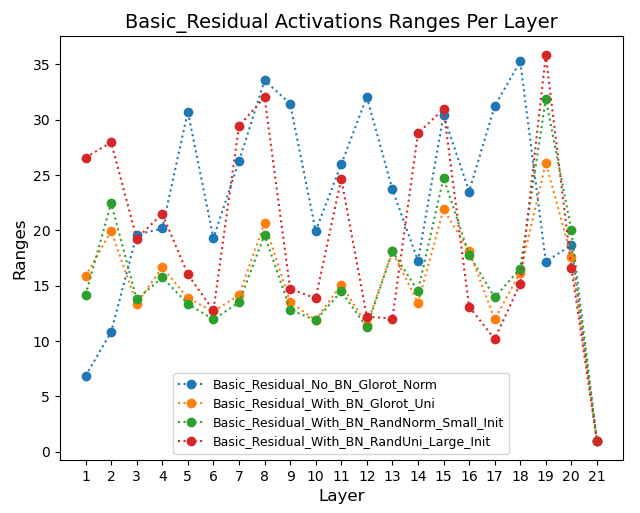} % second figure itself
    \end{minipage}\hfill
    \begin{minipage}{0.5\textwidth}
        \centering
        \includegraphics[width=0.96\textwidth]{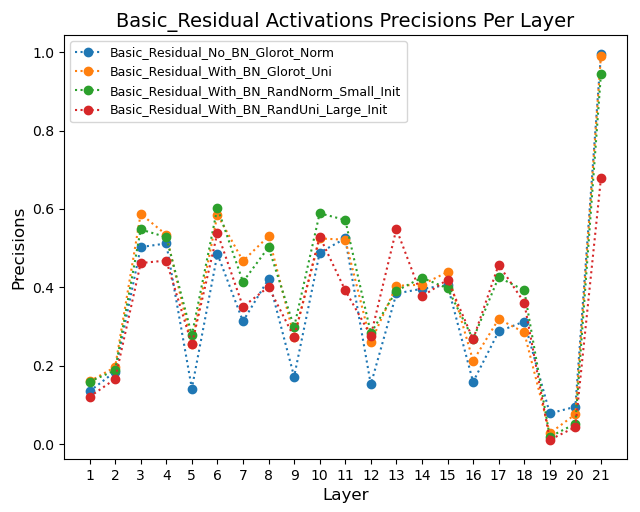} % second figure itself
    \end{minipage}\hfill
    \vspace{-0.07in}
    \caption{\footnotesize{} Layerwise plots for selected Basic-Residual-style networks. All Weights plots are using BN-Folded Weights}
    
\label{fig:basicres-analysis}
\end{figure}

\begin{figure}

    \centering
    \begin{minipage}{0.5\textwidth}
        \centering
        \includegraphics[width=0.98\textwidth]{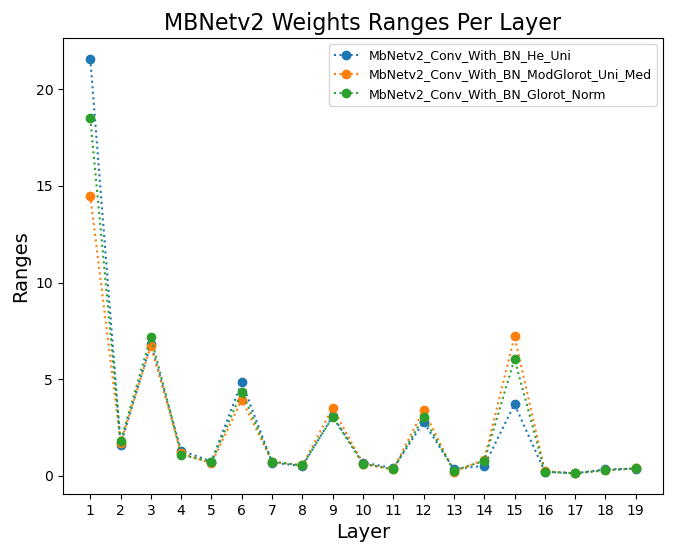} % first figure itself
    \end{minipage}\hfill
    \begin{minipage}{0.5\textwidth}
        \centering
        \includegraphics[width=0.98\textwidth]{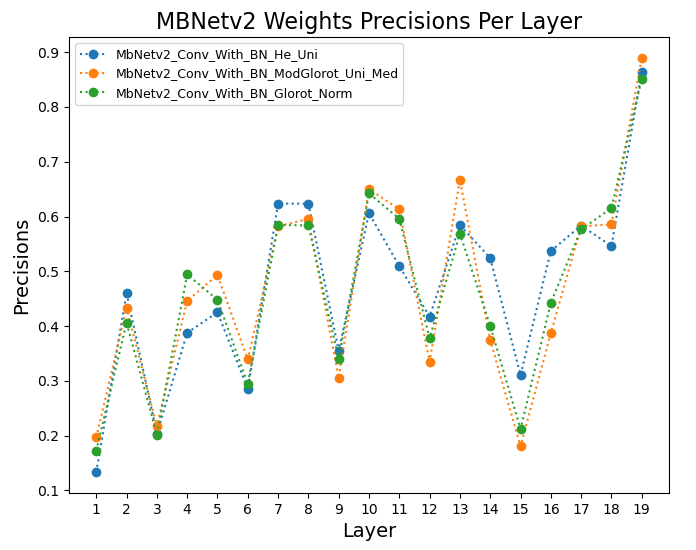} % second figure itself
    \end{minipage}\hfill
    \begin{minipage}{0.5\textwidth}
        \centering
        \includegraphics[width=0.98\textwidth]{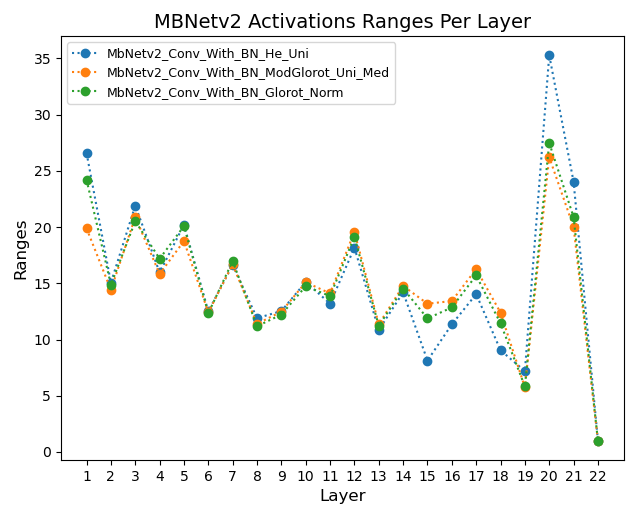} % second figure itself
    \end{minipage}\hfill
    \begin{minipage}{0.5\textwidth}
        \centering
        \includegraphics[width=0.98\textwidth]{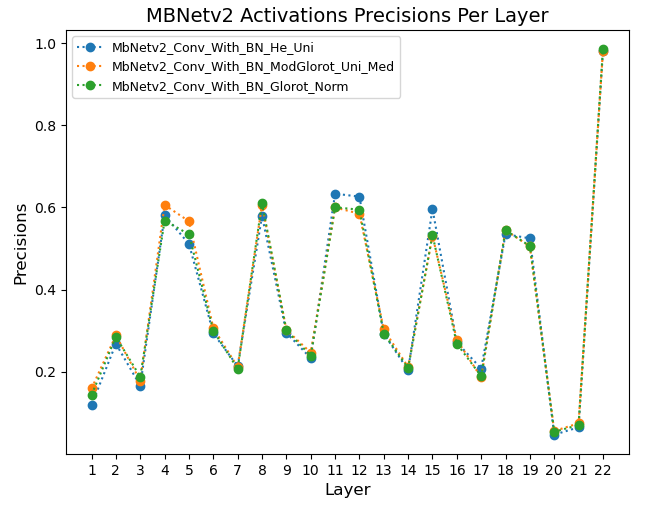} % second figure itself
    \end{minipage}\hfill
    
    \caption{\footnotesize{} Layerwise plots for selected MbNetv2-style networks. All Weights plots are using BN-Folded Weights}
    
\label{fig:mbnetv2-analysis}
\end{figure}

\subsection{Discussion}
\label{random_init_discussion}
We can see in both Tables~\ref{tab:simple_block_result} and~\ref{tab:complex_block_result} that besides affecting the final FP32 accuracy of a given CNN architecture, the weights initialization strategy also has significant impact on the QUINT8 accuracy. Even within the same architecture-type, we can see that the quantized accuracy of networks can vary widely with different initializations.

Particularly worth noting is the markedly improved quantized behaviour in the DWS\_Conv\_With\_BN networks trained using RandUni\_Large initialization and the MbNetv2\_Conv\_With\_BN network initialized with RandUni\_Med distribution. Equally noteworthy is the stark drop in QUINT8 accuracy observed with the DWS\_Conv\_With\_BN networks trained with the HeNorm and HeUni weight initializations, the Regular\_Conv\_With\_BN network trained with ModGlorotUni\_Med initialization, and the Basic\_Residual\_With\_BN network initialized with RandUni\_Large method. As expected, quantized accuracy usually worsened when BatchNorm layers were introduced. This is often attributed to the increased dynamic ranges/distributional shift introduced by BatchNorm Folding.

\begin{figure}
    \centering
    \begin{minipage}{0.48\textwidth}
        \centering
        \includegraphics[width=0.98\textwidth]{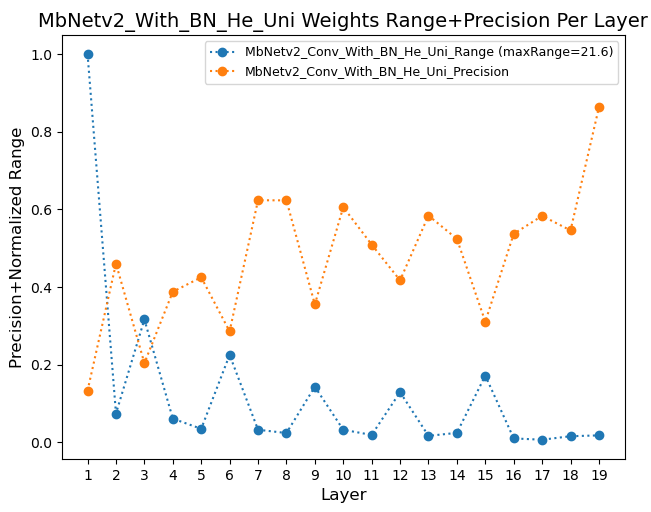} % second figure itself
    \end{minipage}
    \begin{minipage}{0.48\textwidth}
        \centering
        \includegraphics[width=0.99\textwidth]{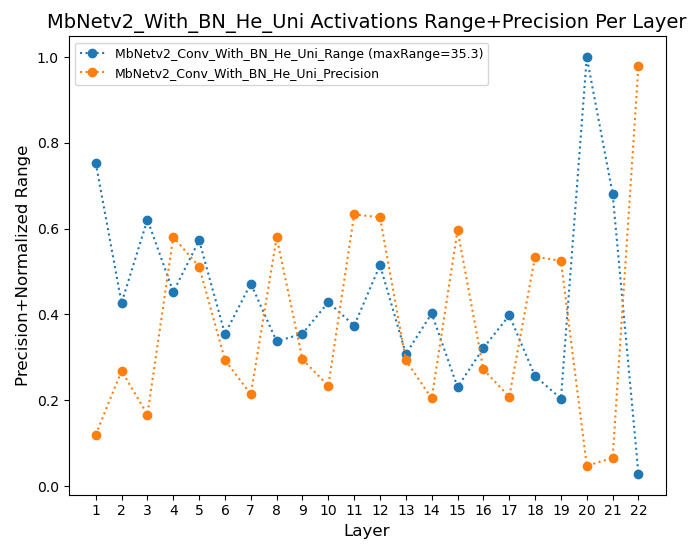} % second figure itself
    \end{minipage}
    \vspace{-0.07in}
    \caption{\footnotesize{} Overlaid plots of range and precision for MbNetv2\_Conv\_With\_BN\_He\_Uni.}
\label{fig:mbnetv2-analysis-2}
\end{figure}

While each CNN architecture is trained on twelve different initialization methods, several initializations led to diverging loss/exploding gradients and consequently, those experiments were ommitted. The normalization introduced by BatchNorm alleviates this issue as expected. To better understand why we are observing the given quantized behaviour, we can use proposed fine-grained analysis proposed in~\cite{stone_masc_thesis} and inspect the distributions of each model layer-by-layer. For example we can notice that models higher quantization accuracy tend to have noticeably lower activation ranges (such as Regular\_Conv\_With\_BN\_GlorotUni in Fig.~\ref{fig:focus-analysis} and Basic\_Residual\_With\_BN\_Glorot\_Uni in Fig.~\ref{fig:basicres-analysis}). Furthermore, models with higher quantization error tend to have significantly higher weights ranges and worse average precision in the early few layers (such as Basic\_Residual\_With\_BN\_RandUni\_Large in Fig.~\ref{fig:basicres-analysis} and MbNetv2\_Conv\_With\_BN\_He\_Uni in Fig.~\ref{fig:mbnetv2-analysis}) indicating a likely loss of important low-level features to quantization noise. Notably in Fig.~\ref{fig:mbnetv2-analysis-2}, we can see how spikes in range often coincide with a drop in precision.

By examining quantization performance across a variety of architecture types, we also see that networks with depthwise separable (DWS) convolution networks tend to have lower average quantization accuracy, highlighting how these networks are more prone to quantization error, and that more tailored initialization would likely improve their performance. Due to the complex interactions of DNNs with quantization, it is hard to pinpoint direct causal relationships between dynamic ranges and quantization robustness. However, large-scale studies appear to reveal trends and patterns that correlate initialization, trained distributions, and quantization robustness. Thus, learning to reason about graphs in a data-driven fashion would be a natural approach to scale this study. Something that Graph Hypernetworks are perfectly suited for. Evidently, there is a large diversity of trajectories that a model can take through the parameter space. Seeing as how initialization can lead to such widely varying learned distributions, it would make sense to investigate how to improve the initialization of models for quantization accuracy, and not just floating-point.

\section{Graph Hypernetworks For Quantization-Robust CNN Initialization}
\label{sec:ghn_intro}
Besides random parameter initialization, GHNs~\cite{GHN1, PPUDA, GHN3} offer an incredibly promising new method for better initialization of DNNs. In contrast to original Hypernetworks~\cite{HyperNetworks}, GHNs can be used for parameter prediction of \textit{unseen} DNN architectures --- that is, networks that the GHN was not trained on --- and thus are much more scalable to new applications. To explore GHNs for resource efficient AI, we focus on adapting GHNs for efficient, quantized DNN parameter prediction as a means to accelerate the design of fixed-precision CNN models for edge deployment. As we have seen how parameter initialization can significantly affect quantized accuracy, we would like to adapt GHNs for more quantization-robust initialization of CNNs. Through learning on a diverse set of efficient, quantized CNN architectures, the graph representational power of GHNs should be useful for predicting quantization-robust parameters for unseen networks and reduce the number of cycles/iterations spent on training quantized models. While we primarily investigate uniform quantization, we show that we can easily quantize the CNNs with different bitwidths. As an extension, we should also be able to model other scalar quantization methods. Thus, GHN-Q becomes a powerful tool for quantization-aware design of efficient CNN architectures.

\begin{figure}
\centerline{\includegraphics[width=9cm]{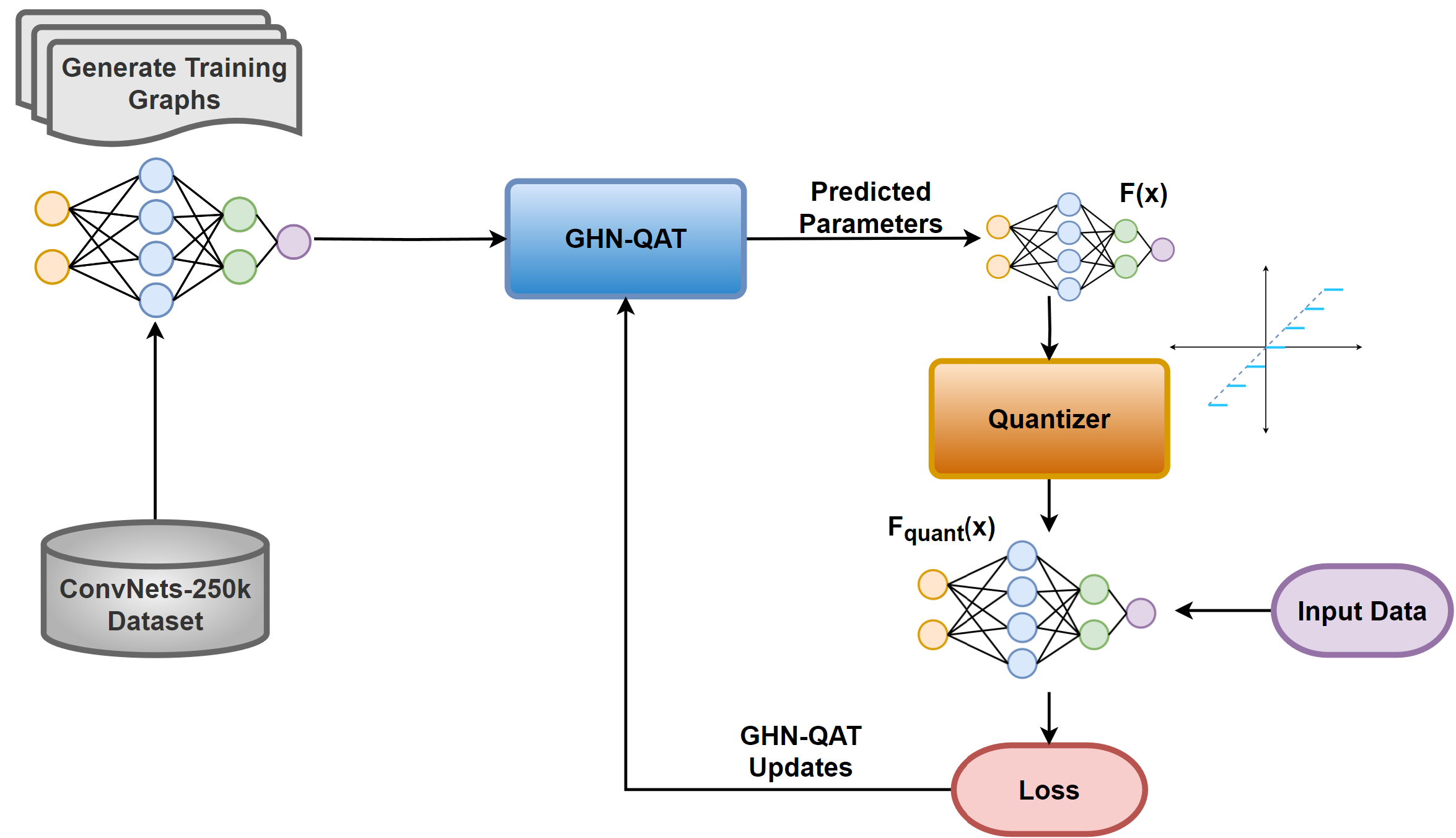}}
\caption[ghnq\_training]{GHN-QAT finetuned on ConvNets-250K. We generate a large number of CNN graphs which are then quantized to target bitwidth for training. Once trained, GHN-QAT can predict robust parameters for unseen CNNs.}
\label{fig:GHNQ_train}
\end{figure}

For this study, we first explore the quantization robustness of GHN-predicted parameters after running float32-precision finetuning on the new efficient-CNN design space. Next, we perform quantization-aware training of the GHN wherein the GHN model is still floating point precision, but the graphs for which parameters are predicted are quantized CNNs. Thus, we are able to study how quantization-robust GHN-predicted parameters are out-of-the-box and compare that to the improvements gained from specifically optimizing for an efficient, quantized neural network graph dataset. Figure~\ref{fig:GHNQ_train} shows how GHN-QAT is finetuned to predict efficient, quantization-robust CNNs.

\subsection{Graph Dataset Generation}
\label{sec:graph_dataset}
For finetuning GHN-2, we generate a dataset of $2.5\times10^5$ mobile-friendly architectures that we call ConvNets-250K. We consider efficient CNN architectures since we are targeting the design of efficient, quantized on-device CNNs. Following a similar procedure to the authors in~\cite{PPUDA}, we randomly generate graphs with various ops and connectivity patterns using the basic building blocks defined in the DARTS design space~\cite{DARTS}. That is, each network has a stem, a random sequence of normal and reduction cells (number of cells randomly varies in a defined range), and then finally a classification head. To generate a random block, we randomly sample connections and operations (including dimensionality/channels of the operation) similar to those described in~\cite{PPUDA, DARTS}, but with a reduced set of operations. The DARTS design space consists of several commonly used blocks such as convolution layers and pooling layers. We have a similar design space but restrict the types of sampled operations to be those that can be efficiently implemented on mobile devices including pooling operations (max and average), convolution\footnote{Kernel sizes of $3\times3$, $5\times5$, and $7\times7$} (separable, regular, and dilated) and finally skip connections. BatchNorm layers are also randomly sampled since their latency can be hidden during mobile inference by folding the op into the preceding conv layer. In future works, more mobile-friendly versions of transformer layers such as those from MobileBERT~\cite{mobileBERT} or Vision Transformer (ViT)~\cite{ViT} could also be explored.

For testing, we generate additional testing sets of out-of-distribution (OOD) network architectures that would be fairly different from our training set such as much deeper networks (much larger number of cells, referred to as ``Deep'' test set), much wider networks (much more channels, ``Wide'' test set), and also DNNs with no BatchNorm layers at all (``BN-Free'' test set). To limit our graph dataset to efficient CNNs, we simply limit number of trainable parameters to a maximum of $10^{7}$ and resample if a given architecture exceeds the limit. A more hardware-aware dataset could involve limiting FLOPs or limiting the search space further to architectures that a given hardware platform is known to perform well with. For example in~\cite{FBNet}, the authors found that optimal graphs searched for Samsung Galaxy S8 were biased towards $5\times5$ depthwise convolution layers while Apple iPhone X preferred $3\times3$ depthwise convolution. Thus certain types of ops, max channel-depth, max kernel size etc. are additional constraints that could be imposed if one wanted to have more hardware-specific optimization. These randomly generated graphs serve as our efficient-CNN graph training set for finetuning GHN. In addition, this could also serve as a search space if we were to use GHN-Q and GHN-QAT for quantized neural architecture search (NAS).

\subsection{Experiment Design}
\label{sec:ghn_experiment_design}
We finetuned a CIFAR-10, DeepNets-1M~\cite{PPUDA} pretrained GHN-2 model obtained from~\cite{PPUDA_Github} on ConvNets-250K graph dataset. Finetuning was run for 100 epochs using CIFAR-10~\cite{cifar10}. Initial learning rate was $0.001$ and reduced by a factor of $0.1$ at epoch 75. GHN-Q is trained with Adam optimizer using $\beta_1 = 0.9$, $\beta_2 = 0.999$, weight decay of $10^{-5}$, training batch-size of 32, and meta-batch-size of 4. We continue to use the weight-tiling and parameter normalization described in~\cite{PPUDA} and we use $s^{(max)}=10$ as the maximum shortest path for virtual edges. The GHN is finetuned from GHN-2 which is a gated-RNN with forward/backward iterations of $N=1$. As we use the ImageNet-pretrained GHN-2 provided by authors of~\cite{PPUDA} at~\cite{PPUDA_Github}, we refer to~\cite{PPUDA} for further architectural details of GHN.

We follow a testing procedure similar to~\cite{PPUDA} and evaluate the trained GHN-Q by comparing the mean CIFAR-10 test accuracy at full, float32 precision to test accuracy at various limited integer precisions (W8/A8, W4/A4, W4/A8, W2/A2 where `W' indicates weight bitwidth and `A' indicates activation bitwidth). As mentioned in~\ref{sec:graph_dataset}, we generate multiple testing sets with different distributions of architectural characteristics. BN-Free networks have no BatchNorm layers. Wide/Deep indicate much wider/deeper nets than training. For handling BatchNorm, we use a test-batch-size of 64 to get batch statistics. We also have to recompute BatchNorm-folded (BN-Fold) weights each batch before quantizing the BN-Fold weights like in~\cite{TFQuantize}. See Eq.~\ref{eq:bn_fold} for exact folding procedure. Quantization encodings use the absolute tensor ranges.

It is also worth noting that we only quantize the weights and activations with a $Quantize()$ operator (using tensorwise, asymmetric, uniform quantization throughout) instead of running fully fixed-point inference. Rounding and truncation errors of fully fixed-point arithmetic will lead to some additional error and are usually implementation-specific (e.g., a table-lookup vs. polynomial approximation-based implementation for activation functions). However, as most of the quantization noise is due to weights and activations, the simulated quantization generally correlates well with on-device accuracy~\cite{TFQuantize, TFQuantWhitePaper, AIMET}.

\subsection{GHN-Q: Predicted Parameters As Quantization-Robust Initialization}
\label{sec:ghn_q_experiment}
We would first like to evaluate whether full precision floating point training on a target design space can train GHN-Q to predict high-performant CNN parameters that are robust to 8-bit uniform quantization. A couple aspects of the method in~\cite{PPUDA} suggest that the parameters predicted by GHN-Q should be compact and quantization-friendly, namely the channel-wise weight tiling and differentiable parameter normalization. Table~\ref{tab:ghnq_results} shows  the results on our different testing splits.

\begin{table*}[h]
    \caption{Testing GHN-Q on unseen quantized networks. CIFAR-10 top-1 and top-5 test accuracy of quantized CNNs are compared to their Float32 accuracy. Presented as (Mean$\%\pm$standard error of mean; Max$\%$). \textbf{ID} indicates in-distribution graphs sampled the same way as training set. \textbf{OOD} are out-of-distribution graphs with characteristics very different from those sampled in training.}
    
    \centering
    \begin{tabular}{c| c| ccc}
    \toprule
     \textbf{Top-1 Accuracy}& \textbf{ID} & \multicolumn{3}{c}{\textbf{OOD}}\\% \vline\\
     \textbf{by Bitwidth} & Test & Deep &  Wide &  BN-Free \\
     \midrule
      Float32 & $71.1\pm0.3; 80.2$ & $68.1\pm0.7; 79.8$ & $69.8\pm0.5; 79.0$ & $37.8\pm1.3; 56.1$ \\
      W8/A8 & $70.9\pm0.3; 80.1$ & $67.9\pm0.7; 79.5$ & $69.6\pm0.5; 78.6$ & $37.5\pm1.3; 56.4$ \\
      W4/A8 & $47.4\pm 0.4; 63.9$ & $39.0\pm 0.7; 62.9$ & $43.8\pm 0.6; 63.3$ & $25.7\pm 0.9; 43.7$\\
      W4/A4 & $37.2\pm0.3; 52.6$ & $30.7\pm0.5; 50.2$ & $34.7\pm0.5; 50.7$ & $21.5\pm0.8; 36.7$\\
      W2/A2 & $11.2\pm0.1; 18.8$ & $10.4\pm0.0; 13.0$ & $11.4\pm0.1; 17.7$ & $11.5\pm0.2; 18.1$\\
      \bottomrule
      \textbf{Top-5 Accuracy} & \textbf{ID} & \multicolumn{3}{c}{\textbf{OOD}}\\% \vline\\
     \textbf{by Bitwidth} & Test & Deep &  Wide &  BN-Free \\
     \midrule
      Float32 & $97.3\pm0.1; 99.0$ & $96.4\pm0.2; 98.8$ & $97.2\pm0.1; 98.7$ & $84.3\pm1.2; 94.8$ \\
      W8/A8 & $97.3\pm0.1; 98.9$ & $96.4\pm0.2; 98.8$ & $97.2\pm0.1; 98.7$ & $84.1\pm1.2; 94.9$ \\
      W4/A8 & $89.8\pm 0.2; 95.9$ & $85.2\pm 0.4; 95.8$ & $88.0\pm 0.3; 95.5$ & $74.6\pm 1.2; 89.8$\\
      W4/A4 & $83.7\pm0.2; 91.7$ & $78.6\pm0.4; 91.0$ & $82.1\pm0.3; 91.4$ & $69.3\pm1.1; 87.0$\\
      W2/A2 & $52.4\pm0.1; 63.8$ & $50.9\pm0.1; 56.6$ & $52.7\pm0.2; 69.6$ & $53.4\pm0.4; 68.9$\\
      \bottomrule
    \end{tabular}
    \label{tab:ghnq_results}
\end{table*}

The parameters predicted by GHN-Q are surprisingly robust despite not having been trained on any kind of quantization. It is particularly interesting that even for 4-bit quantization, the average test accuracy is significantly better than random chance. A likely explanation is that the channel-wise weight tiling and differentiable parameter normalization lead to layerwise distributions that are compact and quantization-friendly. In~\cite{MobileNetsQuantizePoorly}, the authors find that a mismatch between channelwise distributions can lead to significant accuracy loss for post-training quantization (PTQ). The weight tiling method in~\cite{PPUDA} copies predicted parameters across channels and thus minimizes such distributional mismatch by-construction. Additionally, parameter normalization could help produce less heavy-tailed distributions. A detailed analysis of the distributions of GHN-Q predicted parameters would yield a clearer picture. As depthwise-separable convolution is particularly susceptible to distributional mismatch, it would be interesting to test the 8-bit quantized performance of GHN-Q on a test set consisting solely of MobileNet-like CNNs such as those in~\cite{MobileNetV1, MobileNetV2}.

Besides analyzing mean quantized accuracy, quantization robustness needs to be quantified on a per-network basis. An analysis of the mean accuracy change and quantization error (e.g., quantized mean squared error) of individual networks would provide better insight. It would be interesting to see how quantization error may change after 8-bit quantized GHN-Q finetuning even if accuracy remains similar.

In~\cite{PPUDA} there are questions of how well GHN-2 predicted parameters can be used for finetuning on the source task. However, if GHN-Q could be adapted such that predicted parameters can directly start quantization-aware training, there would be significant savings in training CNNs for quantization.

\subsection{GHN-QAT: Finetuning GHNs On Quantized Graphs For Predicting Quantization Friendly Parameters}
\label{sec:ghn_qat_experiment}

While Section~\ref{sec:ghn_q_experiment} investigates the inherent quantization-robustness of GHN-predicted parameters after float32 finetuning, we would expect further gains in quantized accuracy to be made after some form of quantization-aware training (either SimQuant or NoiseQuant). We now consider the use of GHN-QAT to further adapt GHNs specifically for quantization-aware training to predict performant parameters for quantized CNN graphs. More specifically, we simulate quantization (SimQuant) in sampled CNNs and train on the loss arising from quantized inference. In this way, GHN-QAT adapts to the quantization errors induced by quantizing GHN-predicted models (see Fig.~\ref{fig:GHNQ_train}). By finetuning GHNs on a mobile-friendly, \textbf{quantized} CNN architecture space, GHNs learn representations specifically for efficient quantized CNNs.

\begin{table*}[h]
    \caption{Testing GHN-QAT on unseen quantized networks. CIFAR-10 test accuracy of quantized CNNs following bitwidth-specific QAT. Presented as (Mean$\%\pm$standard error of mean; Max$\%$).}
    
    \centering
    \begin{tabular}{c| c| ccc}
    \toprule
     \textbf{Top-1 Accuracy} & \textbf{ID} & \multicolumn{3}{c}{\textbf{OOD}}\\% \vline\\
     \textbf{by Bitwidth} & Test & Deep &  Wide &  BN-Free \\
     \midrule
      W8/A8 & $71.2\pm0.4; 80.9$ & $68.0\pm0.7; 79.7$ & $69.9\pm0.6; 79.6$ & $36.9\pm1.4; 55.1$\\
      W4/A8 & $60.2\pm0.4; 75.3$ & $56.5\pm0.7; 72.5$ & $58.1\pm0.7; 73.5$ & $30.7\pm1.0; 49.2$ \\
      W4/A4 & $52.5\pm0.4; 65.7$ & $50.3\pm0.6; 63.7$ & $50.6\pm0.6; 64.1$ & $24.8\pm0.7; 37.2$ \\
      W2/A2 & $26.3\pm0.3; 37.4$ & $23.1\pm0.3; 34.0$ & $24.6\pm0.4; 35.6$ & $10.6\pm0.1; 13.7$\\
    
    \toprule
     \textbf{Top-5 Accuracy} & \textbf{ID} & \multicolumn{3}{c}{\textbf{OOD}}\\% \vline\\
     \textbf{by Bitwidth} & Test & Deep &  Wide &  BN-Free \\
     \midrule
      W8/A8 & $97.4\pm0.1; 99.0$ & $96.4\pm0.2; 98.8$ & $97.1\pm 0.2; 98.8$ & $84.2\pm1.2; 95.2$\\
      W4/A8 & $94.5\pm0.1; 98.1$ & $92.8\pm0.3; 97.7$ & $94.0\pm0.2; 97.8$ & $80.6\pm1.1; 93.6$ \\
      W4/A4 & $91.9\pm0.1; 96.1$ & $90.5\pm0.3; 95.9$ & $91.3\pm0.2; 95.7$ & $76.5\pm0.9; 89.2$ \\
      W2/A2 & $73.2\pm0.3; 85.3$ & $69.7\pm0.4; 81.4$ & $71.6\pm0.5; 83.9$ & $52.4\pm0.3; 60.9$\\
      \bottomrule
    \end{tabular}
    \label{tab:ghnqat_results}
    \vspace{-0.1in}
\end{table*}

We investigate SimQuant based quantization training (commonly referred to as quantization-aware training/QAT, as described in~\ref{sec:model_quant_err}) on a target design space for limited precision quantization using the same bitwidth settings as in Section~\ref{sec:ghn_q_experiment}. Using SimQuant for W2/A2 proved to be unstable and we found that modelling quantization as uniform noise (NoiseQuant, see Eq.~\ref{eq:quant_noise}) led to much better results. The reported W2/A2 results are from training with NoiseQuant where the sampling distribution is computed based on 2-bit precision. In all cases, GHN-QAT training is precision/bitwidth-specific. Encoding bitwidth into the CNN graph could potentially remove the need for bit-width specific finetuning. Table~\ref{tab:ghnqat_results} shows the top-1 and top-5 accuracy results on different testing splits. To establish the benefits of QAT, we can compare to results in Table~\ref{tab:ghnq_results}. Notably, W2/A2 is now better-than-random accuracy and W8/A8 performance is just as good as Float32 after GHN-QAT.

\subsubsection{Discussion}
\label{sec:discussion}

As demonstrated, we can easily simulate quantization of CNNs to arbitrary precision. Thus, GHN-QAT becomes a powerful tool for quantization-aware design of efficient CNN architectures. The parameters predicted by GHN-QAT are remarkably robust and the QAT finetuning results (see Table~\ref{tab:ghnqat_results}) show a significant improvement over simple full-precision float32 finetuning from Section~\ref{sec:ghn_q_experiment}. This shows a clear benefit to adapting GHNs specifically to predict parameters for quantization-aware graphs. Additional possibilities/challenges of leveraging quantization-aware training, such as learned quantization thresholds or reducing QAT oscillations like in~\cite{nagel_qat_oscillations}, should be explored to further improve GHN-QAT, especially for ``extreme'' low bitwidths. It's possible that such improvements to QAT would make SimQuant more stable for 2-bit quantization. 

From GHN-QAT, we can see that introducing quantization into our GHN training allows for greater use of GHNs for quantization-specific neural network parameter prediction. Thus, demonstrating the potential of using GHN-QAT models for quantization-robust weight initialization of CNNs intended for PTQ. Furthermore, if GHN-QAT-predicted parameters can be used as initialization for directly starting quantization-aware training rather than first training models to convergence in full float precision and then additional QAT, then the training time of quantized models would be significantly reduced.

Besides leveraging GHN-QAT for quantized versions of floating point operations, we should be able to encode quantization information such as bit-width and quantization scheme into the graphs. If used as a form of quantized accuracy prediction, GHN-QAT could also be used in NAS for quantized DNNs and greatly accelerate the process of searching for accurate, quantized CNNs.

\subsubsection{GHN Training and Inference Time}
All experiments were run on an Nvidia RTX 2080 Ti. GHN inference is on an Intel i7-5930K CPU. For each experiment of 100 epochs, the total time taken to train a GHN is around 4 days. However, with the significantly improved initialization that can be achieved within just 0.3s of inference on average, prototyping and experimenting with varying architectures can be rapidly accelerated, especially considering the additional overhead of quantization training.

\subsection{Improving Parameter Prediction for BatchNorm-Free CNNs}
Analyzing initial results of GHN-QAT, we notice that GHN-QAT performs very poorly on BN-Free networks. This can possibly be explained by the under-representation of BN-Free graphs in our dataset. For example, when comparing our W4/A4 BN-Free results in Table~\ref{tab:ghnq_results} to those in Table~\ref{tab:ghnqat_results}, there is a very small improvement in quantized accuracy compared to other graph-types and it would appear that QAT did not significantly improve robustness. This is particularly perplexing since the authors in~\cite{MobileNetsQuantizePoorly} as well as our own studies in Section~\ref{sec:random_inits_experiment} find that BatchNorm consistently increases quantization errors. It is possible that BN-Free networks are too far out-of-distribution and that GHN-QAT has overfit to noise induced by BatchNorm layers. To test this, we finetune another GHN model specifically for BatchNorm-Free architectures and evaluate the GHN-predicted parameters on the same BatchNorm-Free test split from the ConvNets-250k dataset. If it is a representation/OOD issue, we would expect this BN-Free-specific GHN-QAT model to perform much better. For BN-Free finetuning, we generate a new dataset of fifty-thousand BN-Free networks that we call BNFree-ConvNets-50K. We repeat the same experiment decribed in Section~\ref{sec:ghn_experiment_design} but finetuning GHN-2 on BNFree-ConvNets-50K. 

\begin{table*}[h]
    \caption{Testing GHN-QAT on quantized BN-Free networks. ``Percentile'' and ``absolute'' refer to the type of clipping done on weights.}
    
    \centering
    \begin{tabular}{c| c| ccc}
    \toprule
     \textbf{Bitwidth} &\textbf{Top-1 Accuracy} & \textbf{Top-5 Accuracy}\\% \vline\\
     \midrule
      Float32 & $65.7\pm0.8; 72.9$ & $96.7\pm0.2; 98.1$\\
      W8/A8 &  $64.5\pm0.7; 71.3$ & $96.5\pm0.3; 97.9$\\
      W4/A8 (percentile) & $58.0\pm0.8; 65.8$ & $94.9\pm0.4; 97.5$\\
      W4/A8 (absolute) & $50.2\pm1.0; 60.9$ & $92.3\pm0.6; 95.9$\\
      W4/A4 & $49.3\pm0.9; 62.0$ & $92.1\pm0.4; 96.1$\\
    
      \bottomrule
    \end{tabular}
    \label{tab:bnfree_results}
    \vspace{-0.1in}
\end{table*}

\subsubsection{Discussion}
From Table~\ref{tab:bnfree_results}, we can see that network-specific optimization of GHN-QAT can lead to significant improvements in accuracy. In fact, at lower precisions than W8/A8, BN-Free networks significantly close the gap in average quantized accuracy with the other testing distributions. Interestingly, for the W4/A4 quantization setting, we found that using absolute tensor ranges led to unstable training and that percentile clipping (top/bottom 1\%) worked much better. Furthermore, the meta-batchsize (i.e., number of graphs sampled per training step) for W4/A4 was increased to 16 and batch-size reduced to 32 (from 64) in efforts to further stabilize training. While these modifications seemed to improve stability, the loss would still often eventually diverge and training would need to be stopped early. In our results, the W4/A4 GHN-QAT testing was done with a checkpoint from Epoch 8 of training and quantized with percentile clipping. The benefits of percentile clipping in the W4/A4 experiment led to us trying percentile clipping for W4/A8 finetuning as well. We would expect that percentile clipping could also improve accuracy of the other W4/A8 testing distributions.

To improve the stability of low-bitwidth training of BN-Free networks, some stronger regularization on the output distributions predicted by GHN-QAT may be needed. Overall, this study seems to confirm our original hypothesis that lagging BN-Free accuracy was due to these types of networks being too far out-of-distribution.  Thus, demonstrating the value of additional finetuning of GHN-QAT for specific architecture types. This could be useful for certain applications where BN-Free networks have become more common such as in super-resolution~\cite{esrgan_2018}. Overall, the improved BN-Free accuracy suggests the value of optimizing GHN-QAT for specific design spaces. Thus, deeper exploration of transformer-like self-attention layers such as in ViT or MobileBERT has great potential

\section{Conclusions and Future Work}
Design of efficient quantized CNNs for mobile is a complex process wherein even the parameter initialization can have deep impacts on the quantization robustness of the final model. We show that for various common CNN architecture designs, different random initialization can lead to varying quantized accuracies and errors. Building on these insights and the works of~\cite{PPUDA, GHN1}, we show that GHNs offer an exciting new direction for amortizing the costs of designing quantization-robust models, being useful for quantizing CNNs to various integer precisions. A quantization-optimized GHN could offer a much better initialization of candidate CNN models which will eventually be quantized. This in turn could lead to better quantized models and also reduce the time and energy needed to produce optimal fixed-point DNN models. As mentioned in~\cite{GHN3}, GHNs could offer a much more viable alternative to costly pretraining. Similary, a strong GHN trained for predicting quantization-robust parameters might offer a much better alternative to the costly iterative process of tuning models for quantized accuracy; a process that not every researcher has the resources and time to afford. Future work includes additional quantization-aware training of CNNs after GHN-QAT predicted parameters, adapting GHN-QAT for mixed-precision quantized CNNs, and also adapting GHNs for a larger variety of tasks, especially since different tasks will have varying tolerance to quantization errors.

\bibliographystyle{IEEEtran}
\bibliography{main}

%%%%%%%%%%%%%%%%%%%%%%%%%%%%%%%%%%%%%%%%%%%%%%%%%%%%%%%%%%%%

\end{document}